\documentclass[11pt]{article}
\usepackage{acl2012}
\usepackage[utf8]{inputenc}
\usepackage{graphicx}
\usepackage{amsmath}
\usepackage[textsize=tiny]{todonotes}

\newcommand\numberthis{\addtocounter{equation}{1}\tag{\theequation}}

\usepackage{amssymb}
\usepackage{multirow}
\usepackage{makecell}

\usepackage[labelfont=bf]{caption}

\usepackage{enumitem}

\title{Represent, Aggregate, and Constrain: A Novel Architecture for Machine Reading from Noisy Sources}
\author{Jason Naradowsky \hspace{1em} Sebastian Riedel\\
  University College London \\
  London, UK \\
  {\tt \{j.narad, s.riedel\}@cs.ucl.ac.uk}}

\date{}

\begin{document}

\maketitle


\begin{abstract}

In order to extract event information from text, a machine reading model must learn to accurately read and interpret the ways in which that information is expressed.
But it must also, as the human reader must, aggregate numerous individual value hypotheses into a single coherent global analysis, applying global constraints which reflect prior knowledge of the domain.  

In this work we focus on the task of extracting plane crash event information from clusters of related news articles whose labels are derived via distant supervision.  Unlike previous machine reading work, we assume that while most target values will occur frequently in most clusters, they may also be missing or incorrect.  

We introduce a novel neural architecture 
to explicitly model the noisy nature of the data and to deal with these aforementioned learning issues.  Our models are trained end-to-end and achieve an improvement of more than 12.1 F$_1$ over previous work, despite using far less linguistic annotation.  
We apply factor graph constraints to promote more coherent event analyses, with belief propagation inference formulated within the transitions of a recurrent neural network. We show this technique additionally improves maximum F$_1$ by up to 2.8 points, resulting in a relative improvement of $50\%$ over the previous state-of-the-art. \end{abstract}

\section{Introduction}

Recent work in the area of machine reading has focused on learning in a scenario with perfect information.
Whether identifying target entities for simple cloze style queries~\cite{hermann2015,miaoYB15}, or reasoning over short passages of artificially generated text~\cite{weston2015}, short stories~\cite{mctest}, or children's stories~\cite{hill2015}, these systems all assume that the corresponding text is \emph{the unique source} of information necessary for answering the query -- one that not only contains the answer, but does not contain misleading or otherwise contradictory information.

For more practical question answering, where an information retrieval (IR) component must first fetch the set of relevant passages, the text sources will be less reliable and this assumption must be discarded.  
Text sources may vary in terms of their integrity (whether or not they are intentionally misleading or unreliable), their accuracy (as in the case of news events, where a truthful but outdated article may contain incorrect information), or their relevance to the query.  These characteristics necessitate not only the creation of high-precision readers, but also the development of effective strategies for aggregating conflicting stories into a single cohesive account of the event.  

Additionally, while many question answering systems are designed to extract a single answer to a single query, a user may wish to understand many aspects of a single entity or event.  In machine reading,  this is akin to pairing each text passage with multiple queries.  Modeling each query as an independent prediction can lead to analyses that are incoherent, motivating the need to model the dependencies between queries.

\begin{figure*}[t]
\centering
\includegraphics[scale=0.76]{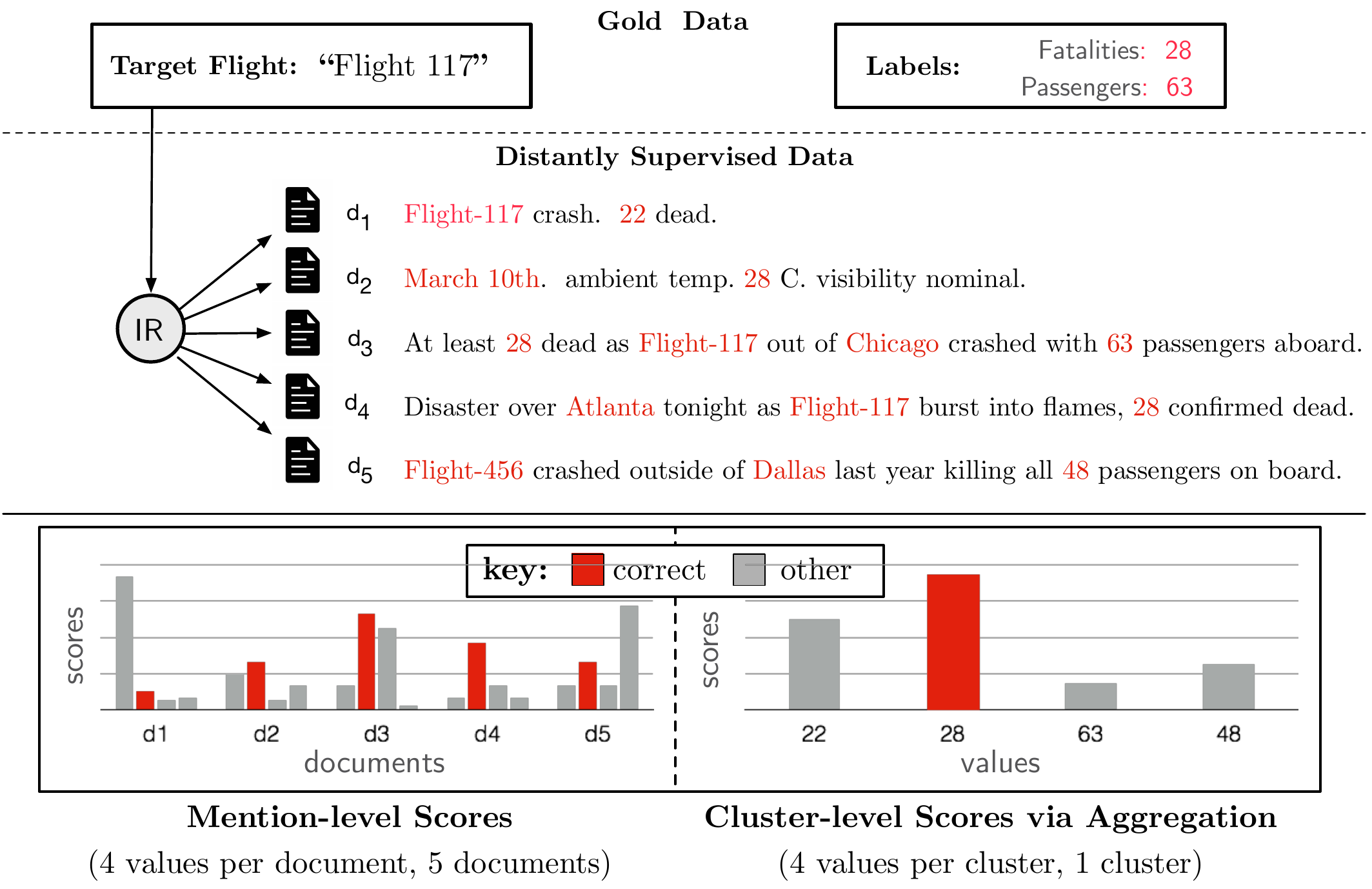}
\caption{An example news cluster.  While we assume all documents mention the target flight, inaccurate information ($d_1$), incorrect labels ($d_2$), and mentions of non-topical events ($d_5$) are frequent sources of noise the model must deal with.  Red tokens indicate mentions of values, i.e. candidate answers.  }
\label{fig:rac_example}
\end{figure*}

We study these problems through the development of a novel machine reading architecture, which we apply to the task of event extraction from news cluster data.  We propose a modular architecture which decomposes the task into three fundamental sub-problems: (1) representation $\&$ scoring, (2) aggregation, and (3) global constraints.  Each corresponds to an exchangeable component of our model.   We explore a number of choices for these components, with our best configuration improving performance by $14.9$ F$_1$, a $50\%$ relative improvement, over the previous state-of-the-art.

\subsection{The Case for Aggregation}

Effective aggregation techniques can be crucial for identifying accurate information from noisy sources.
Figure \ref{fig:rac_example} depicts an example of our problem scenario.  An IR component fetches several documents based on the query, and sample sentences are shown for each document.  The goal is to extract the correct \emph{value}, of which there may be many \emph{mentions} in the text, for each \emph{slot}.  Sentences in $d_1$ express a target slot, the number of fatalities, but the mention corresponds to an incorrect value. This is a common mistake in early news reports.  Documents $d_3$ and $d_4$ also express this slot, and with mentions of the correct value, but with less certainty.

A model which focuses on a single high-scoring mention, at the expense of breadth, will make an incorrect prediction. 
In comparison, a model which learns to correctly accumulate evidence for each value across multiple mentions over the entire cluster can identify the correct information, 
circumventing this problem.  Figure \ref{fig:rac_example} (\emph{bottom}) shows how this pooling of evidence can produce the correct cluster-level prediction.

\section{Model}

In this section we describe the three modeling components of our proposed architecture:
\begin{enumerate}
\item \textbf{Representation and Scoring}, in which a task-specific encoding is generated for each mention, and scored with respect to each slot.
\item \textbf{Aggregation}, in which the scores of each value's mentions are aggregated to produce a single score for each slot-value pair.
\item \textbf{Constraint}, in which we model additional dependencies between pairs of values, and between pairs of slots, to promote more sensible interpretations of the event as a whole.
\end{enumerate}

We begin by defining terminology.  A news cluster $c$ is a set of documents, $\{d_1, ..., d_{|c|}\} \in c$, where each document is described by a sequence of words, $d = (w_1, ..., w_{|d|})$. 
A \emph{mention} is an occurrence of a value in its textual context.  For each value $v \in V$, there are potentially many mentions of $v$ in the cluster, defined as $m \in M(v)$. 
These have been annotated in the data using Stanford CoreNLP~\cite{manning-EtAl:2014:P14-5}.

\subsection{Representations and Scoring}

For each mention $m$ we construct a representation $r(m) \in \mathbb{R}^r$ 
of the mention in its context.  This representation functions as a general ``reading'' or encoding of the mention, irrespective of the type of slots for which it will later be considered.  This differs from some previous machine reading research where the model provides a query-specific reading of the document, or reads the document multiple times when answering a single query~\cite{hermann2015}. As in previous work, an embedding of a mention's context serves as its representation. 

We construct an embedding matrix $E \in \mathbb{R}^{e\times n}$,  using pre-trained word embeddings, where $e$ is the dimensionality of the embeddings and $n$ the number of words in the cluster.  These are held fixed during training.  All mentions are masked and receive the same one-hot vector in place of a pretrained embedding.
From this matrix we embed the context using a two-layer convolutional neural network (CNN), with a detailed discussion of the architecture parameters provided in Section \ref{sec:experiments}.  CNNs have been used in a similar manner for a number of information extraction and classification tasks~ \cite{kim:2014:EMNLP2014,zeng-EtAl:2015:EMNLP} and are capable of producing rich sentence representations~\cite{kalchbrenner-grefenstette-blunsom:2014:P14-1}.\footnote{We also experimented with sequential context embedding models but observed a negligible effect on performance when pursuing a 1-best decoding strategy.}

\subsubsection{Scoring}
Having produced a representation $r(m)$ for each mention $m$, a slot-specific attention mechanism produces $\phi_{\mathrm{mention}}(m,s) \in \mathbb{R}$, representing the compatibility of mention $m$ with slot $s$.  Let $R \in \mathbb{R}^{n \times r}$ be the representation matrix composed of all $r(m)$, and $k(m)$ is the index of $m$ into $R$.  We create a separate embedding, $\pi^s \in \mathbb{R}^{r}$, for each slot $s$, and utilize it to attend over all mentions in the cluster as follows:
\begin{align*} 
u^s &= R\pi^s \numberthis \label{eq:score} \\ 
a^s &= \text{softmax}(u^s) \numberthis \label{eq:attend} \\ 
\phi_{\mathrm{mention}}(m,s) &= a_{k(m)}^s \numberthis \label{eq:mention-score}
\end{align*}
The mention-level scores reflect \emph{an interpretation} of the value's encoding with respect to the slot.  The softmax normalizes the scores over each slot,  supplying the attention, and creating competition between mentions.  This encourages the model to attend over mentions with the most characteristic contexts for each slot.

\subsection{Aggregating Mention-level Scores}
\label{sec:dist-sup}
For values mentioned repeatedly throughout the news cluster, mention scores must be aggregated to produce a single value-level score.  In this section we describe (1) how the right aggregation method can better reflect how the gold labels are applied to the data, 
(2) how domain knowledge can be incorporated into aggregation, and (3) how aggregation can be used as a dynamic approach to identifying missing information.  

\subsubsection{Aggregation as a Model of Distant Supervision
}

In the traditional view of distant supervision \cite{mintz2009}, if a mention is
found in an external knowledge base it is assumed that the mention is an expression of its role in the knowledge base, and it receives the corresponding label.
This assumption does not always hold, and the resulting spurious labels are frequently cited as a source of training noise \cite{riedel2010,hoffmann2011}. 
However, an aggregation over all mention scores provides a more accurate reflection of how distant supervision labels are applied to the data.

If one were to assign a label to each mention and construct a loss using the mention-level scores ($\phi_{\mathrm{mention}}$) directly, it would recreate the hard labeling of the traditional distant supervision training scenario.  Instead, we relax the distant supervision assumption by using a loss on the value-level scores ($\phi_{\mathrm{value}}$), with aggregation to pool beliefs from one to the other.
This explicitly models the way in which cluster-wide labels are applied to mentions, and allows for spuriously labeled mentions to receive lower scores, ``explaining away'' the cluster's label by assigning a higher score to a mention with a more suitable representation.

Two natural choices for this aggregation are max and sum.  Formally, under max aggregation the value-level scores for a value $v$ and slot $s$ are computed as:
\begin{equation}
\label{eq:max-agg}
\phi_{\mathrm{value}}(v,s) = \max_{m\in M(v)} \phi_{\mathrm{mention}}(m,s)
\end{equation}

And under sum aggregation:
\begin{equation}
\label{eq:sum-agg}
\phi_{\mathrm{value}}(v,s) = \sum_{m \in M(v)}\phi_{\mathrm{mention}}(m,s)
\end{equation}

If the most clearly expressed mentions correspond to correct values, max aggregation can be an effective strategy \cite{riedel2010}.  If the data set is noisy with numerous spurious mentions, a sum aggregation favoring values which are expressed both clearly \emph{and frequently} may be the more appropriate choice.

\subsubsection{Weighted Aggregation}
The aforementioned aggregation methods combine mention-level scores \emph{uniformly}, but for many domains one may have prior knowledge regarding which mentions should more heavily contribute to the aggregate score.  
It is straightforward to augment the proposed aggregation methods with a separate weight $\alpha_m$ for each mention $m$ to create, for instance, a weighted sum aggregation:
\begin{equation}
\label{eq:weighted-agg}
\phi_{\mathrm{value}}(v,s) = \sum_{m \in M(v)} \alpha_m \cdot \phi_{\mathrm{mention}}(m,s)
\end{equation}

These weights may be learned from data, or they may be heuristically defined based on a priori beliefs.  Here we present two such heuristic methods.

\paragraph{Topic-based Aggregation}

News articles naturally deviate from the topical event, often including comparisons to related events, and summaries of past incidents.  
Any such instance introduces additional noise into the system, as the contexts of topical and nontopical mention are often similar.  
Weighted aggregation provides a natural foothold for incorporating 
topicality into the model.

We assign aggregation weights heuristically with respect to a simple model of discourse.  We assume every document begins on topic, and remains so until a sentence mentions a nontopical flight number.  This and all successive sentences are considered nontopical, until a sentence reintroduces the topical flight.  
Mentions in topical sentences receive aggregation weights of $\alpha_m = 1.0$, and those in non-topical sentences receive weights of $\alpha_m = 0.0$, removing them from consideration completely.


\paragraph{Date-based Aggregation}

In the aftermath of a momentous event, news outlets scramble to release articles, often at the expense of providing accurate information.  

We hypothesize that the earliest articles in each cluster are the most likely to contain misinformation, which we explore via a measure of information content.  We define the information content of an article as the number of correct values which it mentions. Using this measure, we fit a skewed Gaussian distribution over the ordered news articles, assigning $\alpha_m = ic(d), \forall m \in d$, where $ic(d)$ is the smoothed information content of $d$ as drawn from the Gaussian.

\subsubsection{Known Unknowns}
\label{sec:known-unknowns}
A difficult machine reading problem unique to noisy text sources, where the correct values may not be present in the cluster, is determining whether to predict any value at all.
A common solution for handling such missing values is the use of a threshold, below which the model returns null.  However, even a separate threshold for each slot would not fully capture the nature of the problem.  

Determining whether a value is missing is a trade-off between two factors: (1) how strongly the mention-level scores support a non-null answer, and (2) how much general information regarding that event and that slot is given.  We incorporate both factors by extending the definition of $R$ and its use in Eq.~\ref{eq:score}\textendash Eq.~\ref{eq:mention-score} to include not only mentions, but all words.  Each non-mention word is treated as a mention of the null value:

\begin{equation}
\label{eq:null-agg}
\phi_{\mathrm{value}}(v=null,s) = \sum_{d \in c} \sum_{w \in \{d \setminus M\}} \phi_{\mathrm{mention}}(w,s)
\end{equation}

\noindent where $M$ is the set of mentions.  The resulting null score varies according to both the cluster size and its content.  Smaller clusters with fewer documents require less evidence to predict a non-null value, while larger clusters must accumulate more evidence for a particular candidate or a null value will be proposed instead.  

The exact words contained in the cluster also have an effect.  For example, clusters with numerous mentions of \emph{killed}, \emph{died}, \emph{dead}, will have a higher $\phi_{\mathrm{value}}(v=null,s=${\sc Fatalities}$)$, requiring the model to be more confident in its answer for that slot during training.  Additionally, this provides a mechanism for driving down $\phi_{\mathrm{mention}}(w,s)$ when $w$ is not strongly associated with $s$.

\subsection{Global Constraints}

While the combination of learned representations and aggregation produces an effective system in its own right, 
its predictions may reflect 
a lack of world knowledge.
For instance, we may want to discourage the model from predicting the same value for multiple slots, as this is not a common occurrence.


Following recent work in computer vision which proposes a differentiable interpretation of belief propagation inference ~\cite{ross-cvpr-11,crfasrnn:iccv2015}, we present a recurrent neural network (RNN) which implements inference under this constraint.

\subsubsection{Belief Propagation as an RNN}


\begin{figure*}[t]
\centering
\hspace{-2em}
\includegraphics[scale=0.60]{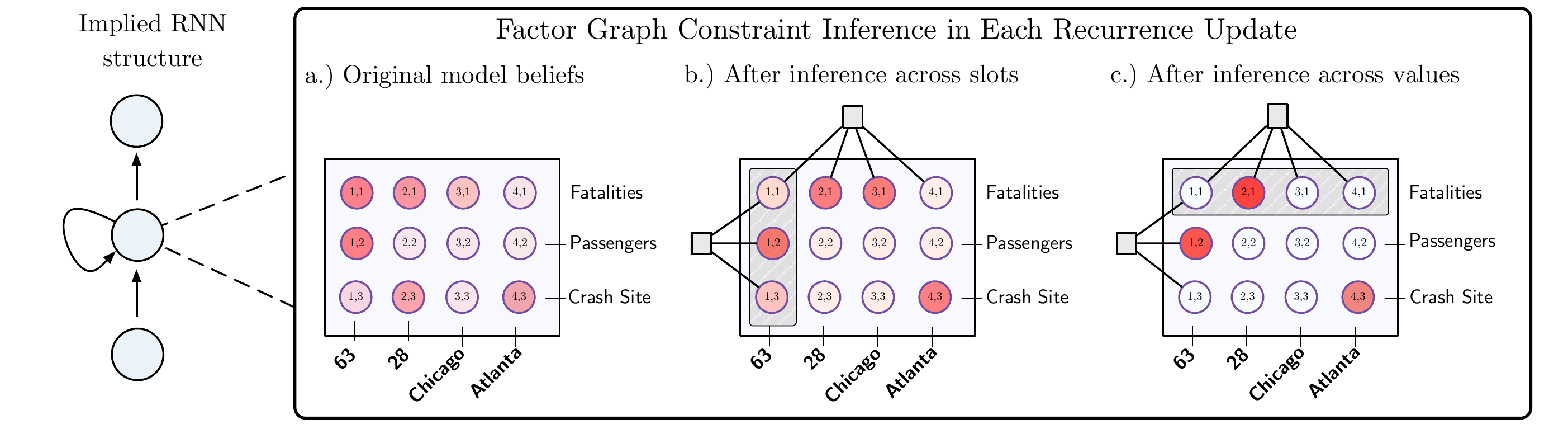}
\caption{Belief propagation constraint inference as an RNN.  Red indicates a true value. 
At each step of LBP inference, the belief of each variable is updated with respect to the two {\sc Exactly-1} factors it is connected to, pushing it closer to discrete values.  In convergence, (c.), values reflect the desired constraint.}
\label{fig:bp_rnn}
\end{figure*}


A factor graph is a graphical model which factorizes the model function using a bipartite graph, consisting of variables and factors.  Variables maintain beliefs over their values, and factors specify scores over configurations of these values for the variables they neighbor.  

We constrain model output by applying a factor graph model to the $\phi_{\mathrm{value}}$ scores it produces.  The slot $s$ taking the value $v$ is represented in the factor graph by a Boolean variable $X_{v,s}$.  Each $X_{v,s}$ is connected to a local factor $u_{v,s}$ whose initial potential is derived from  $\phi_{\mathrm{value}}(v,s)$.
  A combinatorial logic factor,  {\sc Exactly-1}\cite{smith-eisner:2008:EMNLP}, is (1) created for each slot, connected across all values, and (2) created for each value, connected across all slots.  This is illustrated in Figure~\ref{fig:bp_rnn}.  Each {\sc Exactly-1} factor provides a hard constraint over neighboring Boolean variables requiring \emph{exactly one} variable's value to be true, therefore diminishing the possibility of duplicate predictions during inference.


\paragraph{Inference}
The resulting graph contains cycles, preventing the use of exact message passing inference.  We instead treat an RNN as implementing loopy belief propagation (LBP), an iterative approximate message passing inference algorithm.  The hidden state of the RNN is the set of variable beliefs, and each round of message updates corresponds to one iteration of LBP, or one recurrence in the RNN.


There are two types of messages: messages from variables to factors and messages from factors to variables.  The message  that a variable $X$ sends to a factor $f$ (denoted $ \mu_{X \rightarrow f}$) is defined recursively w.r.t. to incoming messages from its neighbors $n(X)$ as follows:

\begin{equation}
\label{eq:standard_v2f}
 \mu_{X \rightarrow f} = \prod_{f' \in n(X) \neq f} \mu_{f' \rightarrow X}
\end{equation}


\noindent and conveys the information ``My other neighbors
jointly suggest I have the posterior distribution $\mu_{v \rightarrow u}(v)$ over my values.''  In our RNN formulation of message passing the initial outgoing message for a variable $X_{v,s}$ to its neighboring {\sc Exactly-1} factors is:
\begin{equation}
\label{eq:bp-setup}
\mu_{X_{v,s} \rightarrow f} = \text{sigmoid}(\phi_{\mathrm{value}}(v, s)) 
\end{equation}

\noindent where the sigmoid moves the scores into probability space.




A message from an {\sc Exactly-1} factor to its neighboring variables is calculated as:

\vspace{-1em}
\begin{align*} 
\neg \mu_{X \rightarrow f} &= 1.0 - \mu_{X \rightarrow f} \numberthis \\ 
Z &= \prod_{X} \neg \mu_{X \rightarrow f} \numberthis \\
\mu_{{\text {\sc Exactly-1}}\rightarrow X} &= \frac{Z}{\neg \mu_{X \rightarrow f}} \numberthis
\end{align*}

All subsequent LBP iterations compute variable messages as in Eq. \ref{eq:standard_v2f}, incorporating the out-going factor beliefs of the previous iteration.

\section{Data}

The Stanford Plane Crash Dataset~\cite{reschke2014} is a small data set consisting of 80 plane crash events, each paired with a set of related news articles.   Of these events, 40 are reserved for training, and 40 for testing, with the average cluster containing more than 2,000  mentions.\footnote{Although it should be noted that only 33 of the training clusters and just 27 of the test clusters contain documents from which to extract information.}  Gold labels for each cluster are derived from Wikipedia infoboxes and cover up to 15 slots, of which 8 are used in evaluation (Figure \ref{tab:per-slot-table}).

We follow the same entity normalization procedure as \newcite{reschke2014}, limit the cluster size to the first 200 documents, and further reduce the number of duplicate documents to prevent biases in aggregation.  We partition out every fifth document from the training set to be used as development data, primarily for use in an early stopping criterion.  We also construct additional clusters from the remaining training documents, and use this to increase the size of the development set.

\section{Experiments}
\label{sec:experiments}

In all experiments we train using adaptive online gradient updates (Adam, see \newcite{kingma2014}).   Model architecture and parameter values were tuned on the development set, and are as follows (chosen values in bold):
\begin{itemize}
\itemsep0em 
\small
\item CNN layer 1 filter width: [3, 5, 8, \textbf{10}]
\item CNN layer 2 filter width: [0, 3, \textbf{5}, 8, 10]
\item CNN layer 1 dim: [5, \textbf{10}, 15, 20]
\item CNN layer 2 dim: [0, 5, \textbf{10}, 15, 20]
\item max pooling [True, \textbf{False}]
\item learning rate: [0.001, \textbf{0.003}, 0.005, 0.01]
\item L2 regularization: [0.001, 0.003, 0.005, \textbf{0.01}]
\item dropout rate: 1-[0.5, 0.6, 0.7, \textbf{0.8}, 0.9, 1.0]
\end{itemize}
The number of training epochs is determined via early stopping with respect to the model performance on development data.  The pre-trained word embeddings are 200-dimensional GLoVe embeddings~\cite{pennington2014glove}.

\subsection{Systems}
\label{sec:baselines}
We evaluate on four categories of architecture:
\paragraph{Existing baselines} \newcite{reschke2014} proposed  several methods for event extraction in this scenario.  We compare against three notable examples drawn from this work: 
\vspace{-0.5em}
\begin{itemize}[leftmargin=*]
\itemsep-0.5em
    \item \textbf{Reschke CRF}: a conditional random field model.
    \item \textbf{Reschke Noisy-OR}: a sequence tagger with a "Noisy-OR" form of aggregation that discourages the model from predicting the same value for multiple slots.
    \item \textbf{Reschke Best}: a sequence tagger using a cost-sensitive classifier, optimized with SEARN~\cite{daume2009}, a learning-to-search framework.
\end{itemize}
Each of these models uses features drawn from dependency trees, local context (unigram/part-of-speech features for up to 5 neighboring words), sentence context (bag-of-word/part-of-speech), words/part-of-speech of words occurring within the value, as well as the entity type of the mention itself.   

\vspace{-0.4em}
\paragraph{Mention-CNN} The representation and scoring components of our architecture, with an additional slot for predicting a null value.  The $\phi_{\mathrm{mention}}(m,s)$ scores are used when constructing the loss and during decoding.  These scores can also be aggregated in a max/sum manner after decoding, but such aggregation is not incorporated during training.\footnote{These models benefit from vastly different training parameters and were trained for 250 iterations with a dropout rate of 0.3.} 

\vspace{-0.4em}
\paragraph{RAC-CNN} Representation, scoring, and aggregation components, trained end-to-end with a cluster-level loss. 
Null values are predicted as described in Sec. \ref{sec:known-unknowns}.

\vspace{-0.4em}
\paragraph{EE-AS Reader} \newcite{kadlec2016} present AS Reader, a state-of-the-art model for cloze-style QA.  Like our architecture, AS Reader aggregates mention-level scores, pooling evidence for each answer candidate. However, in cloze-style QA an entity is often mentioned in complementary contexts throughout the text, but are frequently in similar contexts in news cluster extraction.

We tailor AS Reader to event extraction to illustrate the importance of choosing an aggregation which reflects how the gold labels are applied to the data. EE-AS Reader is implemented by applying Eq.~\ref{eq:score} and Eq.~\ref{eq:attend} to each document, as opposed to clusters, as documents are analogous to sentences in the cloze-style QA task.  We then concatenate the resulting vectors, and apply sum aggregation as before.

\begin{table*}[t]
\centering
\begin{tabular}{l|| c| |r|r|r ||c|c|c || c}
& Aggregation & \multicolumn{3}{c||}{Score} & \multicolumn{3}{c||}{Nulls} & MRR \\
 & &  \textbf{P} & \textbf{R} & \textbf{F$_{1}$} & \textbf{P} & \textbf{R} & \textbf{F$_{1}$} &  \\
    \Xhline{1.2pt}
    \hline
    \hline
  Reschke CRF & -- & 15.9 & 42.5 & 23.2 & -- & -- & -- & -- \\
Reschke Noisy-OR & -- & 18.7 & 37.0 & 24.8 & -- & -- & -- & -- \\
 Reschke Best & -- & 24.5 & 38.6 & 30.0 & -- & -- & -- & --  \\
 \hline
 \multirow{3}{*}{{\sc Mention-CNN}} 
 & None & 11.8 & 25.9 & 15.9 & 55.8 & 76.5 & \textbf{64.6} & -- \\
 & Max & 18.3 & 28.1 & 22.2 & 0.0 & 0.0 & 0.0 & 24.8  \\
 & Sum & 23.6 & 37.8 & 29.0 & 0.0 & 0.0 & 0.0 & 31.1 \\
 \hline
 \multirow{4}{*}{{\sc RAC-CNN}}
 & Max & 21.3 & 28.9 & 24.5 & 41.0 & 26.3 & 32.0 & 26.3  \\
  & Sum & 34.4 & 47.4 & 39.8 & 52.6 & 37.0 & 43.5 & 35.0\\
 \cline{2-9}
 & Date & 34.4 & 47.4 & 39.9 & 51.7 & 37.0 & 43.2 & 35.2\\
 & Topic & 36.6 & 49.6 & 42.1 & 53.2 & 40.7 & 46.6 & 35.7 \\
 \hline
 RAC-CNN-BP-1 & Topic & 38.7 & 53.3 & \textbf{44.9} & 54.6 & 37.0 & 44.1 & \textbf{36.6} \\
 \hline
 EE-AS Reader & Sum & 12.0 & 19.3 & 14.8 & 0.0 & 0.0 & 0.0  & 17.5 \\
 \hline
\end{tabular}
\caption{Event extraction results across different systems on the Stanford Plane Crash Dataset.} 
\label{tab:main-results}
\end{table*}

\subsection{Evaluation}
We evaluate configurations of our proposed architecture across three measures.  The first is a modified version of standard precision, recall, and F$_1$, as proposed by \newcite{reschke2014}.  It deviates from the standard protocol by (1) awarding full recall for any slot when a single predicted value is contained in the gold slot, (2) only penalizing slots for which there are findable gold values in the text, and (3) limiting candidate values to the set of entities proposed by the Stanford NER system and included in the data set release.  Eight of the fifteen slots are used in evaluation.  Similarly, the second evaluation measure we present is standard precision, recall, and F$_1$, specifically for null values.

We also evaluate the models using mean reciprocal rank (MRR).  When calculating the F$_1$-based evaluation measure we decode the model by taking the single highest-scoring value for each slot.  However, this does not necessarily reflect the quality of the overall value ranking produced by the model.  For this reason we include MRR, defined as:
\begin{equation}
    MRR = \frac{1}{|Q|}\sum_{i=1}^{|Q|} \frac{1}{\text{rank}_i}
\end{equation}

\noindent where rank$_i$ is the rank position of the first correct value for a given cluster and slot pair $i$, and $|Q|$, the number of such pairs, is $|C| \cdot |S|$, the product of the total number of clusters with the total number of predicted slots.


\subsection{Results}
Results are presented in Table \ref{tab:main-results}.  In comparison to previous work, our any configuration of our RAC architecture with sum-based aggregation outperforms the best existing systems by a minimum of 9.8 F$_1$.  In comparison to the various {\sc Mention-CNN} systems, it is clear that this improvement is not a result of different features or the use of pre-trained word embeddings, or even the representational power of the CNN-based embeddings.  Rather, it is attributable to training end-to-end with aggregation and a cluster-level loss function.  

\paragraph{Aggregation Results}
With respect to aggregation, sum-based methods consistently outperform their max counterparts, indicating that exploiting the redundancy of information in news clusters is beneficial to the task.  The topic-based aggregation is statistically significant improvement over standard sum aggregation (p $\le 0.0215$), and produces the highest performing unconstrained system.  

Date-based aggregation did not yield a statistically significant improvement over sum aggregation.  We hypothesize that the method is sound, 
but accurate datelines could only be extracted for 31 $\%$ documents.  We did not modify the aggregation weights ($\alpha(m) = 1.0$) for the remaining documents, minimizing the effect of this approach.

The {\sc EE-AS Reader} has the lowest overall performance, which one can attribute to pooling evidence in a manner that is poorly suited to this problem domain.  By placing a softmax over each document's beliefs, what is an advantage in the cloze-style QA setting here becomes a liability: 
the model is forced to predict a value for every slot, for every each document, even when few are truly mentioned.  

\subsection{Effects of Global Constraints}
\begin{figure}   
 \centering
  \hspace{-1em}
\includegraphics[scale=0.22]{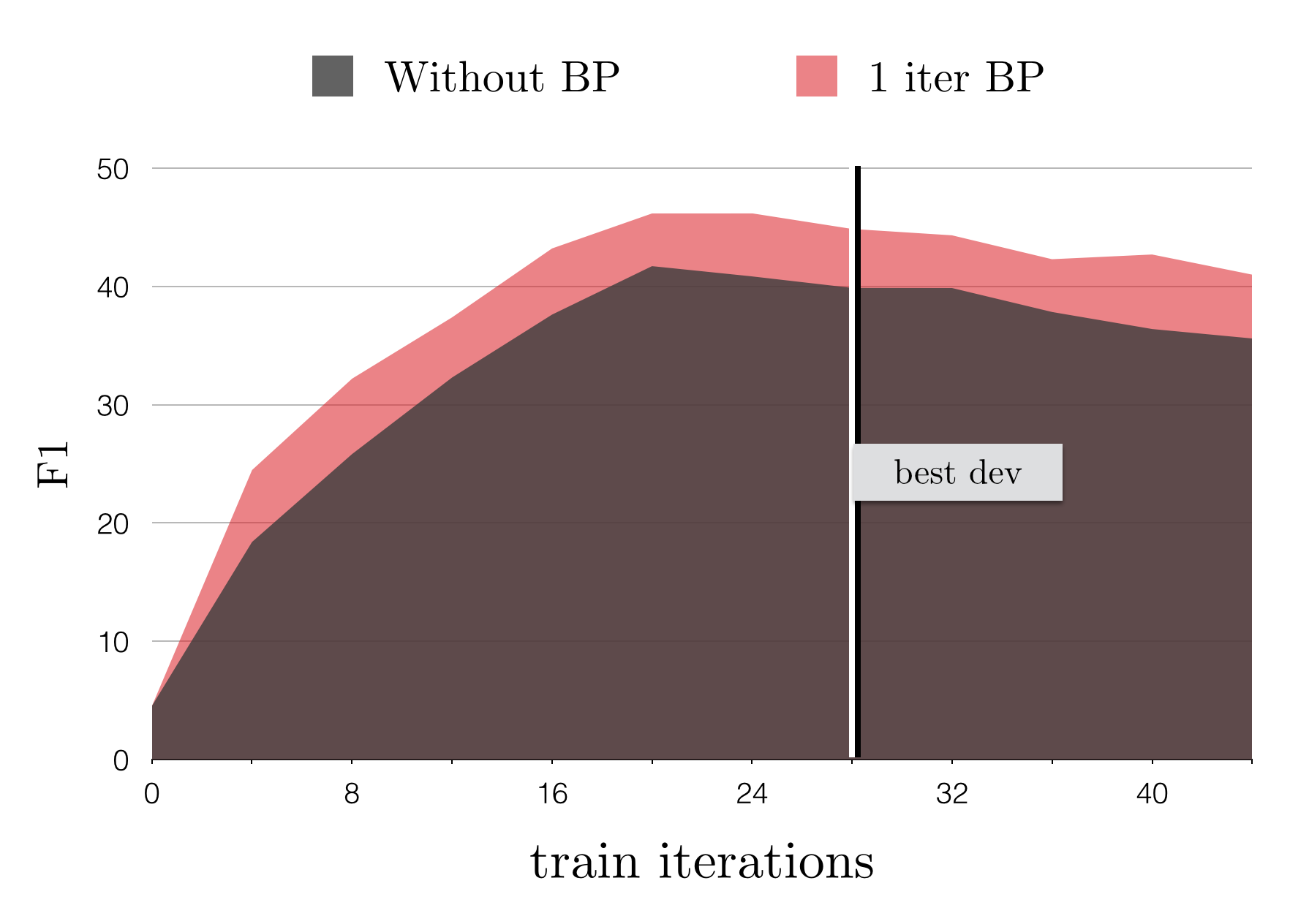}
    \caption{Improvement of BP constraint inference across training iterations}
    \label{fig:bp-plot}
\end{figure}

In Table \ref{tab:constraint-results-table} we show the results of incorporating factor graph constraints into our best-performing system. Performing one iteration of LBP inference produces our highest performance, an F$_1$ of 44.9.  This is 14.9 points higher than Reschke's best system, and a statistically significant improvement over the unconstrained model (p $\le 0.0313$).  The improvements persist throughout training, as shown in Figure \ref{fig:bp-plot}.

Additional iterations reduce performance. This effect is largely due to the constraint assumption not holding absolutely in the data.  For instance, multiple slots can have the null value, and zero is common value for a number of slots.  
Running the constraint inference for a single iteration encourages a 1-to-1 mapping from values to slots, but it does not prohibit it.  This result also implies that a hard heuristic decoding constraint time would not be as effective. 

\begin{table}[t]
\centering
\begin{tabular}{c|| l|l|l || l|l|l}
  & \multicolumn{3}{c||}{Score} & \multicolumn{3}{c}{Nulls}\\
  \textbf{BP}  & \textbf{P} & \textbf{R} & \textbf{F$_{1}$ }  & \textbf{P} & \textbf{R} & \textbf{F$_{1}$ } \\
    \Xhline{1.2pt}
    \hline
    \hline
0 & 36.6 & 49.6 & 42.1 & 53.2 & 40.7 & \textbf{46.2} \\
1 & 38.7 & 53.3 & \textbf{44.9} & 54.6 & 37.0 & 44.1 \\
2 & 37.2 & 50.4 & 42.8 & 55.0 & 40.7 & 46.8 \\
conv. & 37.2 & 50.4 & 42.8 & 54.1 & 40.7 & 46.5 \\
 \hline
\end{tabular}
\caption{Results using global constraints.}
\label{tab:constraint-results-table}
\end{table}

\subsection{Error Analysis}
We randomly selected 15 development set instances which our best model predicts incorrectly.  Of these, we find three were incorrectly labeled in the gold data as errors from the distance supervision hypothesis (i.e., ``zero chance'' being labeled for 0 survivors, when the number of survivors was not mentioned in the cluster), and should not be predicted. 

Six were clearly expressed and should be predictable, with highly correlated keywords present in the context window, but were assigned low scores by the model.  
We belief a richer representation which combines the generalization of CNNs with the discrete signal of n-gram features \cite{lei-barzilay-jaakkola:2015:EMNLP} may solve some of these issues.

Four of the remaining errors appear to be due to aggregation errors.  Specifically, the occurrence of a particular punctuation mark with far greater than average frequency resulted in it being predicted for three slots.  While these could be filtered out, a more general solution may be to build a representation based on the actual mention (``\emph{Ryanair}''), in addition to its context. This may reduce the scores of these mentions to such an extent that they are removed from consideration.

Table \ref{tab:per-slot-table} shows the accuracy of the model on each slot type. The model is struggles with predicting the {\sc Injuries} and {\sc Survivors} slots.  The nature of news media leads these slots to be discussed less frequently, with their mentions often embedded more deeply in the document, or expressed textually.  For instance, it is common to express $s$={\sc Survivors}, $v=0$ as ``no survivors'', but it is impossible to predict a 0 value in this case, under the current problem definition.

\section{Related Work}

\paragraph{Multi-document and Paraphrase-driven IE}
Our work is thematically similar to work in multi-document information extraction~\cite{Mann:2005:MIE:1219840.1219900} and summarization~\cite{Barzilay:1999:IFC:1034678.1034760}, where the content of many input documents is unified into a cohesive understanding.  However, in addition to the many modeling choices we propose, the data sets used in existing work were not linked to specific events. The same denoising nature of these tasks has clear implications for automated fact-checking ~\cite{vlachos-riedel:2014:W14-25}, but no comparable models currently exist.

In contrast, the the IDEST system of \newcite{krause2015idest} is an example of previous work which uses automatically constructed clusters of news articles in order to train an event embedding model.  However, the focus of IDEST is to improve event clustering, not information extraction, which is reflected in its comparatively simple and heuristically-driven extraction method.

\begin{table}[t]
\centering
\begin{tabular}{r| c|c}
\textbf{Slot} & \textbf{correct} & \textbf{findable} \\
    \Xhline{1.2pt}
Aircraft Type & 10 & 18 \\
Crash Site & 15 & 19 \\
Crew & 13 & 21 \\
Fatalities & 10 & 18 \\
Injuries & 0 & 9 \\
Operator & 15 & 18 \\
Passengers & 10 & 16 \\
Survivors &  0& 16
\end{tabular}
\caption{Per-slot accuracies of our best model}
\label{tab:per-slot-table}
\end{table}


\paragraph{Attention and Aggregation in Machine Reading}
In terms of reading methodology, our scoring method is a slot-specific interpretation of the attentive reader~\cite{hermann2015}, and our sum aggregation is closely-related to \newcite{kadlec2016}, with differences described previously in Sec.~\ref{sec:baselines}.   A similar method is found in the entailment model of \newcite{parikh2016}, where alignment scores (between a premise and a hypothesis) are generated via attention and summed.  Recent machine reading models have used an iterative attention to refine model predictions ~\cite{DBLP:journals/corr/SordoniBB16}.  Such methods play a role similar to our factor graph constraint, though they incorporate no prior knowledge.

\paragraph{Extensions to Distant Supervision}
Aggregation in our framework is a means to weaken strong distant supervision assumptions, and, unlike \newcite{mintz2009}, it does not assume that each mention-level occurrence of a value must express the given relation.  In this respect, it exists as a fully-differentiable analog to the work of \newcite{hoffmann2011}, and closely related to the ``expressed at least once'' constraint of \newcite{riedel2010}.  Both allow the model to ignore mislabeled instances in certain circumstances.  Recently, ~\newcite{lin-EtAl:2016:P16-1} have also proposed the use of neural mechanisms to reduce the effect of mislabeled instances, using attention to select the most useful sentences for relation extraction, as we use attention to select the most informative mentions.

\subsection{Connections to Pointer Networks}

A pointer network uses a softmax to normalize a vector the size of the input, to create an output distribution over the dictionary of inputs~\cite{vinyals2015nips}.  This assumes that the input vector is the size of the dictionary, and that each occurrence is scored independently of others.  If an element appears repeatedly throughout the input, each occurrence is in competition not only with other elements, but also with its duplicates.

Here the scoring and aggregation steps of our proposed architecture can together be viewed as a pointer network where there is a redundancy in the input which respects an underlying grouping.  Here the softmax normalizes the scores over the input vector, and the aggregation step again yields an output distribution over the dictionary of the input.

\section{Conclusion and Future Work}
In this work we present a machine reading architecture designed to effectively read collections of documents in noisy, less controlled scenarios where information may be missing or inaccurate.  Through attention-based mention scoring, cluster-wide aggregation of these scores, and global constraints to discourage unlikely solutions, we improve upon the state-of-the-art on this task by 14.9~F$_1$.

In future work, the groundwork laid here may be applied to larger data sets, and may help motivate the development of such data.  Larger noisy data sets would enable the differentiable constraints and weighted aggregation to be included during the optimization, and tuned with respect to data.  
In addition, we find the incorporation of graphical model inference into  neural architectures to be a powerful new tool, and potentially an important step towards incorporating higher-level reasoning and prior knowledge into neural models of NLP.

\bibliographystyle{acl2012}
\bibliography{event_reader}

\end{document}